\documentclass[11pt]{article}
\usepackage{geometry}
\usepackage{coling2020}
\usepackage{times}
\usepackage{url}
\usepackage{latexsym}
\usepackage{microtype}

\colingfinalcopy 

\title{LT3 at SemEval-2020 Task 9: Cross-lingual Embeddings for Sentiment Analysis of Hinglish Social Media Text}

\author{Pranaydeep Singh and Els Lefever \\
LT3, Language and Translation Technology Team\\
Department of Translation, Interpreting and Communication -- Ghent University\\
Groot-Brittanni\"{e}laan 45, 9000 Ghent, Belgium  \\
{\tt firstname.lastname@ugent.be} 
}

\date{}

\begin{document}
\maketitle
\begin{abstract}
This paper describes our contribution to the SemEval-2020 Task 9 on Sentiment Analysis for Code-mixed Social Media Text. We investigated two approaches to solve the task of Hinglish sentiment analysis. The first approach uses cross-lingual embeddings resulting from projecting Hinglish and pre-trained English FastText word embeddings in the same space. The second approach incorporates pre-trained English embeddings that are incrementally retrained with a set of Hinglish tweets.
The results show that the second approach performs best, with an F1-score of 70.52\% on the held-out test data.

\end{abstract}

\section{Introduction}

\blfootnote{
    \hspace{-0.45cm}  
    This work is licensed under a Creative Commons 
    Attribution 4.0 International License.
    License details:
    \url{http://creativecommons.org/licenses/by/4.0/}.
}

The emergence of Web 2.0 has allowed people to easily share their opinion on a variety of topics. Whereas in the past companies and policy makers used to conduct surveys to know the opinion of people on certain products, services or policies, they now have access to a wide range of easily accessible data to gather the public’s sentiment~\cite{Liu2012}. 

To automatically derive opinions from text,  researchers have designed the task of sentiment analysis (SA), which deals with ``the computational study of opinions, sentiments and emotions expressed in text"~\cite{Kumar2012}. An important challenge when applying sentiment analysis to user-generated data is caused by code-mixing and non-standard language use. In linguistics, code-mixing traditionally refers to the embedding of linguistic units (phrases, words, morphemes) into an utterance of another language~\cite{Myers-scotton1993}. 
The phenomenon of code-mixing frequently occurs in spoken languages, such as the combination of English with Hindi (so-called ``Hinglish”), or English with Spanish (so-called ``Spanglish"). More recently, code-mixing is increasingly being used in written text as well, as non-native English speakers often combine English with their mother tongue when using social media. In the case of Hinglish, an additional challenge is added because people do not only mix languages, but also use English phonetic typing to write Hindi words instead of using the Devanagari script. 

In order to investigate Sentiment Analysis for Code-mixed Social Media Text, 
Patwa et al.~\shortcite{patwa2020sentimix} have organized the SentiMix task, which consists in predicting the sentiment of a given code-mixed tweet. The sentiment labels are positive, negative, or neutral, and the code-mixed languages are English-Hindi and English-Spanish. Besides the sentiment labels, the authors also provide language tags at the word level, being \emph{en} (English), \emph{spa} (Spanish), \emph{hi} (Hindi), \emph{mixed}, and \emph{univ} (e.g., symbols, @ mentions, hashtags). This paper presents our research performed for the ``Hinglish" (English-Hindi) sentiment analysis subtask of SemEval-2020 Task 9.

The remainder of this paper is structured as follows. In Section~\ref{related}, we provide an overview of the related research. Section~\ref{data} describes the data used to train and evaluate the system. Section~\ref{system_descr} introduces the two approaches developed to perform Hinglish sentiment analysis, while  Section~\ref{results} discusses the results obtained for the task. Section~\ref{conclusion} concludes this paper.

\section{Related Research}\label{related}

A first line of research for \textbf{sentiment analysis} applies supervised machine learning approaches~\cite{Joshi2010,Vanhee2017}. These approaches, however, require large amounts of labeled data, which are often lacking for low(er)-resourced languages.
Another important line of research uses machine translation systems to (1) map subjectivity lexicons to other languages~\cite{Mihalcea2007,Meng2012} or to (2) transfer sentiment information from a high-resource source language to a low-resource target language~\cite{Banea2008}. Rasooli et al.~\shortcite{Rasooli2018} use annotation projection to project supervised labels from the source languages to the target language and a direct transfer approach to develop SA systems. 

More recently, researchers have started to investigate cross-lingual embeddings for the task of SA. The idea of these embeddings stems from the idea of Mikolov et al.~\shortcite{Mikolov2013} that vector spaces in different languages share a certain similarity. By creating monolingual spaces and then learning a projection from one language to another, there is no need for large parallel corpora. Mikolov et al.~\shortcite{Mikolov2013} learn a linear mapping from one space to another and optimize the performance by using the most common words from both languages and by using a bilingual lexicon. As large bilingual lexicons are often not available, there was a need to either completely eliminate or drastically reduce the size of the required bilingual lexicon. To address this issue, Artetxe et al.~\shortcite{artetxe2017} propose a very simple self-learning
approach that exploits the structural similarity of embedding spaces, and works with as little bilingual evidence as a 25 word dictionary or even an automatically generated list of numerals. Research by Barnes et al.~\shortcite{Barnes2018} attempts to learn bilingual sentiment embeddings, which jointly train the projection and the sentiment component to represent sentiment information in the source and target language. Their method uses a bilingual lexicon, an annotated sentiment corpus in the source language and monolingual embeddings for the source and target language. Their experimental results show the need for a dedicated high-quality sentiment lexicon in order to achieve a satisfactory performance. More recently, transformer-based approaches~\cite{Conneau2018} have been used for cross-lingual knowledge transfer. These approaches, however, require significant pretraining and a lot of low-resource languages are not accounted for in the pretrained models.

Applying \textbf{sentiment analysis to code-mixed social media data}, however, offers a number of challenges for standard NLP approaches. These approaches are usually trained on large monolingual corpora (e.g.~English or Hindi), and not on mixed data. In addition, social media language is characterized by informal language use (abbreviations, spelling mistakes, flooding, emojis, etc.), which causes a considerable drop in performance for standard NLP approaches that are trained on standard data~\cite{Ritter2011}.
Related research on computational models for code-mixing is scarce because of the lack of large code-mixed resources, which makes it hard to apply data-greedy approaches. 
Seminal work in sentiment analysis (SA) of Hindi text was done by \cite{Joshi2016}, who introduce a Hindi-English code-mixed dataset for sentiment analysis and propose a system to SA that learns sub-word level representations in LSTM instead of character- or word-level representations. Pratapa et al.~\shortcite{Pratapa2018} compare three bilingual word embedding approaches to perform code-mixed sentiment analysis and Part-of-Speech tagging. Their results show that the applied bilingual embeddings do not perform well, and that multilingual embeddings might be a better solution to process code-mixed text. This is mainly because code-mixed text contains particular semantic and syntactic structures that do not occur in the respective monolingual corpora. 
Recently, there is a lot of attention for NLP approaches on code-mixed data, as illustrated by the ``Fourth Workshop on Computational Approaches to Linguistic Code-switching”\footnote{\texttt{https://code-switching.github.io/2020/}}.

In the proposed research, we experimented with two different approaches to tackle sentiment analysis for Hinglish: (1) an approach using cross-lingual embeddings resulting from projecting Hinglish and English embeddings in the same space, and (2) an approach incorporating pre-trained English embeddings that are incrementally retrained with Hinglish information.

\section{Data}\label{data}
\subsection{Task Data}
The task data consists of 15,131 instances of Hinglish tweets. Each tweet has a sentiment tag (positive, negative, neutral), and every token in the tweet is tagged with a language label: \emph{en} (English), \emph{hi} (Hindi), \emph{mixed} and \emph{univ} (e.g.~symbols, @ mentions, hashtags). Since the data consists of transliterated Hindi words from an informal source like social media, there is an abundance of non-standard spellings, omission of characters and flooding, all of which add to the challenge of understanding this text. Although the task organizers provide a language label for every token, we opted to omit this information for our experiments. This way, the task would better represent a real-world problem where no language labels are available.
\subsection{Additional Data}
In addition to the data provided for the shared task, we decided to collect a set of Hinglish Tweets as a supervision source for creating better representations for Hinglish words. These tweets are not annotated for sentiment, and were directly scraped from the Twitter API. Since the API does not classify Hinglish as a separate language, 252,183 Hindi tweets were scraped, and subsequently tweets with Devanagari characters were removed, resulting in a set of 138,589 Hinglish tweets. 

\section{System Description}\label{system_descr}

Hinglish is an amalgamation of English and transliterated Hindi. However, since resources on code-mixed Hindi are very limited, we have to find alternative ways to obtain supervision for understanding code-mixed Hindi text and ideally combine the information with already available resources for English. We approached the task of analysing Hinglish code-mixed text from two different angles:

\begin{enumerate}
    \item Hinglish as an independent third language, not inheriting from Hindi or English
    \item Hinglish as an extension of English, with an extended vocabulary

\end{enumerate}

\subsection{Hinglish as an Independent Language (H-IND)}
For our first approach, we treat the scraped set of 138,589 Hinglish Tweets as a corpus of monolingual Hinglish data, and train FastText~\cite{Bojanowski2017} word embeddings for this corpus. We opted for FastText because it is fast, efficient and also accounts for sub-word information which could be crucial in this context. Contextualized word-embedding methods like BERT~\cite{Devlin2019} and ELMo~\cite{Peters2018}, although more advanced, are not ideal for this particular task as they typically require more information. \\
To make the model more robust and perceptive to English words which were not present in our original Twitter corpus, we  also incorporate pre-trained English FastText word embeddings trained on the vast Common Crawl Corpus\footnote{https://commoncrawl.org/}. Since the two sets of embeddings are in separate $n$-dimensional spaces, they need to be projected in a shared space. For the projection, we resort to the methods presented by Artexte et al.~\shortcite{artetxe-etal-2018-robust}, using similarity distributions between the embeddings to create a small artificial bilingual dictionary, which is then used for alignment while also being improved iteratively. The code\footnote{https://github.com/artetxem/vecmap} for the alignment process was made available by the authors. We used the Seed Dict method with default parameters for the most part, except for the CSLS Neighborhood of 8 to define the SeedDict, and a 15,000 cutoff to define the initial vocabulary. Unit norm was used to normalize the embeddings. After obtaining joint cross-lingual embeddings for English and Hinglish, we proceed  with the task of sentiment classification using the data provided for the shared task. The training set of 15,131 tweets was used to train various classifiers, while the validation set of 3,000 tweets was used to tune the parameters of the network. We experimented with a number of standard classifiers incorporating the cross-lingual embeddings: 
\begin{enumerate}
    \item \textbf{Support Vector Machine} (scikit-learn): Linear SVM with L2 penalization, trained with Hinge loss and Regularization Parameter of 1.0;
    \item \textbf{BiLSTM Classifier} (Pytorch): Bi-LSTM encoder followed by a Softmax layer. The size of the hidden layer was 128 and we incorporated 4 layers in our model. This was followed by a single linear layer and the whole system was trained with Cross-Entropy Loss optimized with Stochastic Gradient Descent (SGD) with a lr of 1e-3; 
    \item \textbf{CNN-Based Classifier} (Pytorch): CNN layers with 100 filters each, with kernel sizes ranging from 1 up to 5. The CNN Layers are followed by a Linear layer for classification. The model was penalized with standard Negative Log Likelihood (NLL) Loss and optimized with the Adam optimizer. 
\end{enumerate}
\def\arraystretch{1.5}

\subsection{Hinglish as an Extension (H-EXT)}

The intuition behind the second approach is to simply treat Hinglish words as additional words to the English vocabulary that are missing from the pre-trained embeddings. As a starting point, we use the same FastText pre-trained English embeddings trained on the Common Crawl Corpus (See Section 3.2), and incrementally retrain them with the scraped Hinglish tweets to accomodate new Hinglish words into the vocabulary. As a precaution to make sure that the original English embeddings do not deteriorate due to the incremental pre-training, we freeze the embeddings for the words occurring in the corpus. For classification, the same set of classifiers (and settings) was used as for the experiments described in Section 3.2.

\section{Results and Discussion}\label{results}

As can be seen from Table 1, both systems perform satisfactorily for the task, exceeding the task baseline F1-score of 0.654 by a considerable margin.  It is also worth noting that the CNN-based classifiers work better for this particular task than on the one hand more complicated models like stacked LSTMs, or on the other hand simpler models like Linear SVMs. 

\begin{table}[h!]
\centering
\begin{tabular}{c|c}
             & Avg F1 \\ \hline
baseline    & 0.654 \\ \hline
H-IND SVM    &  0.5947      \\ \hline
H-IND BiLSTM &  0.6689      \\ \hline
H-IND CNN    &   0.6873     \\ \hline
H-EXT SVM    &  0.6194      \\ \hline
H-EXT BiLSTM &  0.6711      \\ \hline
H-EXT CNN    &   \textbf{0.7052}    \\ 
\end{tabular}
\caption{Macro-Averaged F1-scores for all the classifiers for both the H-IND and H-EXT models, as tested on the held-out test data of the competition.}
\end{table}

A more detailed overview of the precision and recall scores for the best performing CNN classifiers is presented in Table 2. 
The H-IND CNN system was our official submission for the task and placed 14th on the final leaderboard (Codalab user: c1pher), while the best team on the leaderboard obtained an Average F1-scrore of 0.75.
It is interesting to note that the H-EXT CNN system outperforms the H-IND CNN system. This is possibly due to the transfer of the embeddings to a shared space in the H-IND system, which deteriorates the quality of the embeddings considerably, whereas incremental re-training appears to be a safer option, since the original English embeddings where frozen. 

\def\arraystretch{1.5}
\begin{table}[h!]
\centering
\begin{tabular}{c|ccc|ccc}
      &           & Val Set &         &           & Test Set &         \\ \cline{2-7} 
      & Precision & Recall  & Avg. F1 & Precision & Recall   & Avg. F1 \\ \hline
H-IND &   0.6712        & 0.6514        &   0.6552      &     0.7081&           0.6836&          0.6873         \\ \hline
H-EXT &   \textbf{0.6860}        & \textbf{0.6598}        &   \textbf{0.6673}      &     \textbf{0.7204}      &  \textbf{0.6988}        & \textbf{0.7052}        \\ \hline
\end{tabular}
\caption{Precision, Recall and Macro-Averaged F1-score for the CNN-based Classifiers for both models H-IND and H-EXT, on the validation (\emph{Val Set}) as well as the test (\emph{Test Set}) data.}
\label{resultsUnsupervised}
\end{table}

An error analysis has shown that there is still a lot of room for improvement. The FastText embeddings are certainly not perfect due to the limited amount of tweets collected. Frequent words like~\textit{ham} (English: we) and \textit{bharat} (English: India) were well represented in the scraped tweets, whereas rarer words like \textit{abhigyaan} (English: knowledge source) and \textit{kanoon} (English: law) had few occurrences, thus diminishing the quality of the FastText embeddings that were trained based on this corpus. 

\section{Conclusion}\label{conclusion}

In this paper we demonstrate that it is possible to create a sentiment analysis system
for Hinglish by a) treating it as an independent language and b) treating it as an extension of English with additional vocabulary. Both models beat the task baseline convincingly. In addition, we achieve these results without using the language labels provided for every word in the task, thus demonstrating that these methods can be employed with real-world data and can be scaled to any code-mixed language in general. 
In future research, it would be interesting to evaluate the performance of these models on other code-mixed tasks. It would also be worthwhile to use contextual embeddings like BERT and XLM, since these methods have significantly outperformed conventional word embeddings in all multilingual NLP tasks.

\bibliographystyle{coling}
\bibliography{lt3}

\begin{thebibliography}{}

\bibitem[\protect\citename{Artetxe \bgroup et al.\egroup }2017]{artetxe2017}
M.~Artetxe, G.~Labaka, and E.~Agirre.
\newblock 2017.
\newblock Learning bilingual word embeddings with (almost) no bilingual data.
\newblock In {\em Proceedings of the 55th Annual Meeting of the Association for
  Computational Linguistics (Volume 1: Long Papers)}, pages 451--462,
  Vancouver, Canada. ACL.

\bibitem[\protect\citename{Artetxe \bgroup et al.\egroup
  }2018]{artetxe-etal-2018-robust}
M.~Artetxe, G.~Labaka, and E.~Agirre.
\newblock 2018.
\newblock A robust self-learning method for fully unsupervised cross-lingual
  mappings of word embeddings.
\newblock In {\em Proceedings of the 56th Annual Meeting of the Association for
  Computational Linguistics (Volume 1: Long Papers)}, pages 789--798,
  Melbourne, Australia, July. Association for Computational Linguistics.

\bibitem[\protect\citename{Banea \bgroup et al.\egroup }2008]{Banea2008}
C.~Banea, R.~Mihalcea, J.~Wiebe, and S.~Hassan.
\newblock 2008.
\newblock {Multilingual subjectivity analysis using machine translation}.
\newblock In {\em Proceedings of the 2008 Conference on Empirical Methods in
  Natural Language Processing (EMNLP 2008)}, pages 127--135.

\bibitem[\protect\citename{Barnes \bgroup et al.\egroup }2018]{Barnes2018}
J.~Barnes, R.~Klinger, and S.~Schulte~im Walde.
\newblock 2018.
\newblock {Bilingual Sentiment Embeddings: Joint Projection of Sentiment Across
  Languages}.
\newblock In {\em Proceedings of the 56th Annual Meeting of the Association for
  Computational Linguistics}, pages 2483--2493, Melbourne, Australia.

\bibitem[\protect\citename{Bojanowski \bgroup et al.\egroup
  }2017]{Bojanowski2017}
P.~Bojanowski, E.~Grave, A.~Joulin, and T.~Mikolov.
\newblock 2017.
\newblock {Enriching Word Vectors with Subword Information}.
\newblock {\em Transactions of the Association for Computational Linguistics},
  5:135--146.

\bibitem[\protect\citename{Conneau \bgroup et al.\egroup }2018]{Conneau2018}
A.~Conneau, G.~Lample, R.~Rinott, A.~Williams, S.~Bowman, H.~Schwenk, and
  V.~Stoyanov.
\newblock 2018.
\newblock {Evaluating crosslingual sentence representation}.
\newblock In {\em Proceedings of the 2018 Conference on Empirical Methods in
  Natural Language Processing (EMNLP 2018)}.

\bibitem[\protect\citename{Devlin \bgroup et al.\egroup }2019]{Devlin2019}
J.~Devlin, M.-W. Chang, K.~Lee, and K.~Toutanova.
\newblock 2019.
\newblock {BERT}: Pre-training of deep bidirectional transformers for language
  understanding.
\newblock In {\em Proceedings of the 2019 Conference of the North {A}merican
  Chapter of the Association for Computational Linguistics: Human Language
  Technologies, Volume 1 (Long and Short Papers)}, pages 4171--4186,
  Minneapolis, Minnesota. Association for Computational Linguistics.

\bibitem[\protect\citename{Joshi \bgroup et al.\egroup }2010]{Joshi2010}
A.~Joshi, A.R. Balamurali, and P.~Bhattacharyya.
\newblock 2010.
\newblock {A fall-back strategy for sentiment analysis in Hindi: a case study}.
\newblock In {\em Proceedings of ICON 2010: 8th International Conference on
  NLP}, India.

\bibitem[\protect\citename{Joshi \bgroup et al.\egroup }2016]{Joshi2016}
A.~Joshi, A.~Prabhu, M.~Shrivastava, and V.~Varma.
\newblock 2016.
\newblock Towards sub-word level compositions for sentiment analysis of
  {H}indi-{E}nglish code mixed text.
\newblock In {\em Proceedings of {COLING} 2016, the 26th International
  Conference on Computational Linguistics}, pages 2482--2491, Osaka, Japan.

\bibitem[\protect\citename{Kumar and Sebastian}2012]{Kumar2012}
A.~Kumar and T.M. Sebastian.
\newblock 2012.
\newblock Sentiment analysis on twitter.
\newblock {\em International Journal of Computer Science Issues},
  9(3):372--378.

\bibitem[\protect\citename{Liu}2012]{Liu2012}
B.~Liu.
\newblock 2012.
\newblock Sentiment analysis and opinion mining.
\newblock {\em Synthesis lectures on human language technologies}, 5(1):1--167.

\bibitem[\protect\citename{Meng \bgroup et al.\egroup }2012]{Meng2012}
X.~Meng, F.~Wei, X.~Liu, M.~Zhou, G.~Xu, and H.~Wang.
\newblock 2012.
\newblock {Cross-lingual mixture model for sentiment classification}.
\newblock In {\em Proceedings of the 50th Annual Meeting of the Association for
  Computational Linguistics}, pages 572--581, Jeju Island, Korea.

\bibitem[\protect\citename{Mihalcea \bgroup et al.\egroup }2007]{Mihalcea2007}
R.~Mihalcea, C.~Banea, and J.~Wiebe.
\newblock 2007.
\newblock {Learning multilingual subjective language via cross-lingual
  projections}.
\newblock In {\em Proceedings of the 45th Annual Meeting of the Association for
  Computational Linguistics}, pages 976--983.

\bibitem[\protect\citename{Mikolov \bgroup et al.\egroup }2013]{Mikolov2013}
T.~Mikolov, K.~Chen, G.~Corrado, and J.~Dean.
\newblock 2013.
\newblock {Efficient estimation of word representations in vector space}.
\newblock In {\em Proceedings of the ICLR Workshop Papers}.

\bibitem[\protect\citename{Myers-Scotton}1993]{Myers-scotton1993}
C.~Myers-Scotton.
\newblock 1993.
\newblock {\em {Dueling Languages: Grammatical Structure in Code-Switching}}.
\newblock Claredon, Oxford.

\bibitem[\protect\citename{Patwa \bgroup et al.\egroup
  }2020]{patwa2020sentimix}
P.~Patwa, G.~Aguilar, S.~Kar, S.~Pandey, S.~PYKL, B.~Gamb{\"a}ck,
  T.~Chakraborty, T.~Solorio, and A.~Das.
\newblock 2020.
\newblock Semeval-2020 task 9: Overview of sentiment analysis of code-mixed
  tweets.
\newblock In {\em Proceedings of the 14th International Workshop on Semantic
  Evaluation ({S}em{E}val-2020)}, Barcelona, Spain, December. Association for
  Computational Linguistics.

\bibitem[\protect\citename{Peters \bgroup et al.\egroup }2018]{Peters2018}
M.~Peters, M.~Neumann, M.~Iyyer, M.~Gardner, C.~Clark, K.~Lee, and
  L.~Zettlemoyer.
\newblock 2018.
\newblock Deep contextualized word representations.
\newblock In {\em Proceedings of the 2018 Conference of the North {A}merican
  Chapter of the Association for Computational Linguistics: Human Language
  Technologies, Volume 1 (Long Papers)}, pages 2227--2237, New Orleans,
  Louisiana, June. Association for Computational Linguistics.

\bibitem[\protect\citename{Pratapa \bgroup et al.\egroup }2018]{Pratapa2018}
A.~Pratapa, M.~Choudhury, and S.~Sitaram.
\newblock 2018.
\newblock Word embeddings for code-mixed language processing.
\newblock In {\em Proceedings of the 2018 Conference on Empirical Methods in
  Natural Language Processing (EMNLP 2018)}, pages 3067--3072. Association for
  Computational Linguistics.

\bibitem[\protect\citename{Rasooli \bgroup et al.\egroup }2018]{Rasooli2018}
M.S. Rasooli, N.~Farra, A.~Radeva, F.~Yu, and K.~McKeown.
\newblock 2018.
\newblock Cross-lingual sentiment transfer with limited resources.
\newblock {\em Machine Translation}, 32(1--2):143--165.

\bibitem[\protect\citename{Ritter \bgroup et al.\egroup }2011]{Ritter2011}
A.~Ritter, S.~Clark, M.~Etzioni, and O.~Etzioni.
\newblock 2011.
\newblock {Named entity recognition in tweets: an experimental study}.
\newblock In {\em Proceedings of the 2011 Conference on Empirical Methods in
  Natural Language Processing (EMNLP 2011)}, pages 1524--1534.

\bibitem[\protect\citename{Van~Hee \bgroup et al.\egroup }2017]{Vanhee2017}
C.~Van~Hee, M.~Van~de Kauter, O.~De~Clercq, E.~Lefever, B.~Desmet, and
  V.~Hoste.
\newblock 2017.
\newblock Noise or music? investigating the usefulness of normalisation for
  robust sentiment analysis on social media data.
\newblock {\em Traitement Automatique des Langues}, 58(1):63--87.

\end{thebibliography}

\end{document}